\pdfoutput=1

\documentclass[11pt]{article}

\usepackage{authblk}
\usepackage[final]{ACL2023}

\usepackage{times}
\usepackage{latexsym}

\usepackage{times}
\usepackage{latexsym}
\usepackage{subfig}
\usepackage{graphicx}
\usepackage{times}
\usepackage{latexsym}
\usepackage{multirow}
\usepackage{amsfonts}
\usepackage{algorithm2e}
\usepackage{multicol}
\usepackage{multirow}
\usepackage{todonotes}
\usepackage{breqn}
\usepackage{amsmath}
\usepackage{arydshln}
\usepackage{mathtools}
\usepackage{color, colortbl}
\usepackage{array, makecell}
\usepackage{xcolor}
\usepackage{todonotes}

\usepackage[T1]{fontenc}
\usepackage{enumitem}

\usepackage[utf8]{inputenc}

\usepackage{microtype}

\usepackage{inconsolata}
\usepackage{pifont}
\usepackage{comment}
\usepackage{url}
\definecolor{LightRed}{rgb}{0.99,0.8,0.8}
\definecolor{LightGreen}{rgb}{0.6,9,0.6}

\newcommand{\task}{\texttt{TeCFaP}}
\newcommand{\newdataset}{\texttt{TEMP-COFAC}}
\newcommand{\taskfull}{\textbf{Te}mporally  \textbf{C}onsistent \textbf{Fa}ctuality \textbf{P}robe}
\newcommand{\llm}{\text{LLMs}}
\newcommand{\llmfull}{\text{Large Language Models}}
\newcommand{\newmodel}{\texttt{CoTSeLF}}
\newcommand{\newmodelfull}{\textbf{Co}nsistent-\textbf{T}ime-\textbf{Se}nsitive \textbf{L}earning \textbf{F}ramework}
\newcommand{\ctsrl}{\texttt{CTSRL}}

\newcommand{\bert}{\texttt{BERT}}
\newcommand{\xlm}{\texttt{RoBERTa}}
\newcommand{\gpt}{\texttt{GPT}}
\newcommand{\gptj}{\texttt{GPT-J}}
\newcommand{\llama}{\texttt{LLaMA}}
\newcommand{\llamathirteen}{\texttt{LLaMA[13B]}}
\newcommand{\llamaseven}{\texttt{LLaMA[7B]}}

\newcommand{\falcon}{\texttt{Falcon}}
\newcommand{\gptfour}{\texttt{GPT-4}}
\newcommand{\tfive}{\texttt{T5}}
\newcommand{\xmark}{\ding{55}}
\newcommand{\llamathirty}{\texttt{LLaMA[30B]}}
\newcommand{\claudethree}{\texttt{Claude-3}}

\title{Temporally Consistent Factuality Probing for Large Language Models} 

\author[1,2]{\textbf{Ashutosh Bajpai}}
\author[1,*]{\textbf{Aaryan Goyal}}
\author[1,*]{\textbf{Atif Anwer}}
\author[1]{\textbf{Tanmoy Chakraborty}}
\affil[1]{Indian Institute of Technology Delhi, India}
\affil[2]{ Wipro Research, India}
\affil[ ]{\textit {\{eez228482, aaryan.goyal.ee120, atif.anwer.ee120,tanchak\}@ee.iitd.ac.in}}

\begin{document}
\maketitle
\def\thefootnote{*}\footnotetext{Equal contribution}
\begin{abstract}

The prolific use of \llmfull\ (\llm) as an alternate knowledge base requires them to be factually consistent, necessitating both correctness and consistency traits for paraphrased queries. Recently, significant attempts have been made to benchmark datasets and metrics to evaluate \llm\ for these traits. However, structural simplicity (subject-relation-object) and contemporary association in their query formulation limit the broader definition of factuality and consistency. In this study, we introduce \task, a novel \taskfull\ task to expand the consistent factuality probe in the temporal dimension. To this end, we propose \newdataset, a high-quality dataset of prefix-style English query paraphrases. Subsequently, we extend the definitions of existing metrics to represent consistent factuality across temporal dimension. We experiment with a diverse set of \llm\ and find most of them performing poorly on \task. Next, we propose a novel solution \newmodel\ (\newmodelfull) combining multi-task instruction tuning (MT-IT) with consistent-time-sensitive reinforcement learning (\ctsrl) to improve temporally consistent factuality in \llm. Our experiments demonstrate the efficacy of \newmodel\ over several baselines.

\end{abstract}

\section{Introduction}
\label{sec:introduction}

Large Language Models (\llm) are pivotal in propelling the advancement of Artificial General Intelligence (AGI) by acquiring self-learning capabilities for complex tasks \cite{ge2023openagi}. A key development within \llm\ is the ability for {\em temporal reasoning} - comprehending, processing, and reasoning about time-related concepts, temporal dependencies, chronological sequences, and the nuanced, consistent temporal relationship of events. This ability is vital for a myriad of domain-specific tasks, including but not limited to summarizing timelines, tracking disease progression (medical), scheduling events (planning), managing contracts (legal), historical analysis (archaeology), and identifying tasks dependencies (project management).

{\textbf{Why is consistent factuality important?}}
Knowledge bases (KBs) were the foremost choice in factual knowledge retrieval tasks before the appearance of \llm. A KB is a structured database containing a collection of facts (subject, relation, object) \cite{lan2021survey}. There has been a surge of interest in using pre-trained \llm\ as KBs. \citet{alkhamissi2022review}  presented an extensive review on significant developments \cite{petroni2019language} \cite{Dhingra_2022} \cite{heinzerling2021language} in this direction. One of the biggest
appeals of using \llm\ as KBs is that a query can be written in natural language instead of relying on a specific KB schema \cite{elazar2021measuring}. Due to the complex nature of language, the semantic meaning can be expressed in multiple surface forms. Accurate and consistent retrievals, despite a change in surface form, are two fundamental traits required from \llm\ to replace KBs. Built on a manually engineered schema that dictates the possible set of entities and relations, KBs ensure factual and consistent answers \cite{alkhamissi2022review}. On the contrary, inconsistent factuality is reported as a widespread problem in \llm, especially in autoregressive setting \cite{tam2022evaluating}. 

{\textbf{Probing consistent factuality.}}
Usually, factuality (accuracy) is used as a widespread metric in probing \llm\ to check linguistic capabilities such as commonsense \cite{zhang-etal-2020-language-embeddings,forbes2019neural} and reasoning \cite{talmor-etal-2020-olmpics,kassner-etal-2020-pretrained}. On the other hand, an increase in public access of \llm\ requires them to consistently respond to user-specific diversities in surface forms of a query. \citet{ravichander-etal-2020-systematicity} measured consistency through paired probes. \citet{elazar2021measuring} extended the work by investigating and improving the consistency of \llm\ behavior across different factual knowledge types. 

{\textbf{Our novel task and dataset.}} So far, the consistent factuality probing in \llm\ has been predominantly contemporary -- the existence of a subject and an object associated via a relation is coeval, i.e., a typical query from PARAREL dataset \cite{elazar2021measuring} is -- \textit{X was born in Y} in which  subject \textit{X} ({\it person}) and object \textit{Y} ({\it location}) are connected via relation \textit{born-in}. Note that the subject and object are contemporaneous; both exist simultaneously. As information keeps on getting generated, maintained and lost over time, the above formulation is insufficient in capturing the temporal association between the queried entity (subject) and expected value entity (object), therefore providing a poor representation of the overall consistent factuality of \llm.

\begin{figure}[!t]
\centering
\includegraphics[width=0.90\columnwidth]{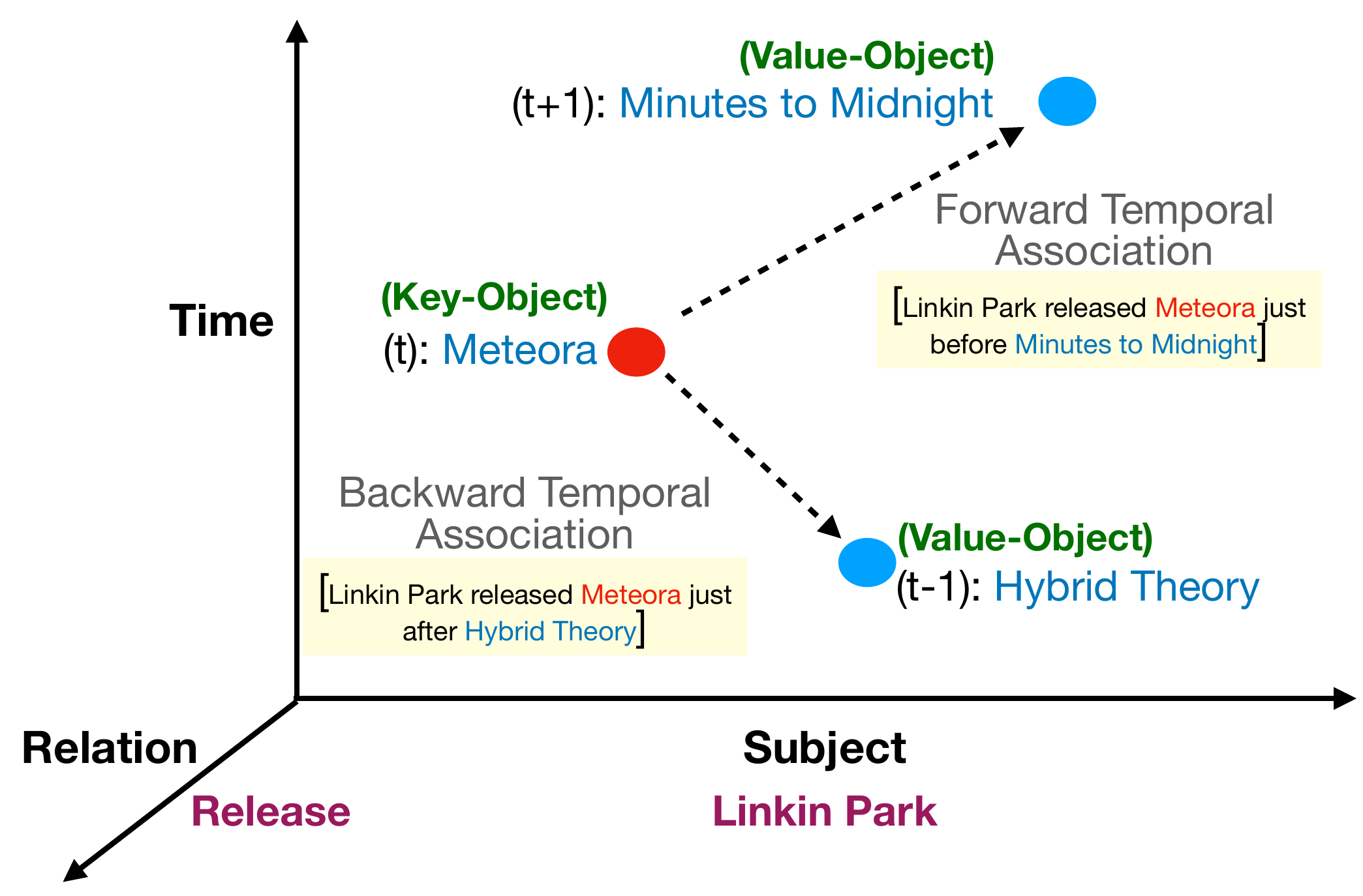}
\caption{Symbolic representation of the \task\  objective. An entity \textit{key_object} holds a temporal relationship with another entity \textit{value_object} via a \textit{subject-relation} pair in either direction -- forward or backward.}
\label{fig:taskintution}
\vspace{-5mm}
\end{figure}

To address this, we present \taskfull\ (\task), a novel task accompanied by a new dataset, \newdataset. \task\ seeks to exploit the temporal association among entities via a \textit{subject-relation} pair to represent temporally consistent factuality of \llm. It defines the query structure in the form of (\textit{key\_object}, \textit{subject-relation}, \textit{value\_object}). The proposed formulation expands the probe in the temporal dimension. As we observe in Figure \ref{fig:taskintution}, the space is three-dimensional -- \textit{subject}, \textit{relation}, and \textit{time}. Since time has a directional attribute, we expect temporal association to be either in a forward or backward direction. For example, the query 
\textit{Hybrid Theory was released by linkin park just before [Meteora]} is a {\em forward} direction probe where a \textit{key\_object} (Hybrid Theory) placed at time \textit{t} is temporally associated with a \textit{value\_object} (Meteora) in time-space at $(t+1)$ via a subject (\textit{Linkin Park}) and a relation (\textit{release-by}). At present, our probe is limited to only strict associations where the expected \textit{value_object} is either located at $(t+1)$ or $(t-1)$ step in temporal space w.r.t \textit{key_object} at time $t$. Inspired by \citet{timeml}, strict association is achieved via trigger words such as \textit{immediately before}, \textit{right after}, \textit{soon after}, etc. We observe the destitute performance of \llm\ on \task\ metrics (defined in Section \ref{sec:task_framework}) -- temporal factuality, temporal consistency, and temporally consistent factuality are in [$0.95\%$ - $3.63\%$], [$13.28\%$ - $64.87\%$] and [$0\%$ to $2\%$] range, respectively in zero-shot setting. 

{\textbf{Our proposed model.}} We present \newmodel, a \newmodelfull, built on a multi-task instruction-tuned (MT-IT) framework followed by consistent-time-sensitive reinforcement learning (\ctsrl) to improve temporally consistent factuality in \llm. We compare \newmodel\ with several baselines and show it outperforming the best baseline  \cite{tan-etal-2023-towards} by $12.7\%$, $10.9\%$ and $90.4\%$, respectively, for temporal factuality, temporal consistency, and temporally consistent factuality.

{\textbf{Contributions.}} In short, we make the following contributions through this study\footnote{Source code and dataset are available at \url{https://github.com/ab-iitd/tecfap}}:
\begin{itemize}
[noitemsep,nolistsep,topsep=0pt,leftmargin=1em]
    \item We establish the need for temporally consistent factuality in \llm\ and propose \task, a novel task (Section \ref{sec:task_framework}).
    \item We create \newdataset, a novel dataset consisting of $66$ diverse \textit{subject-relation} pairs and $8$ paraphrase samples each for forward and backward temporal association for a given \textit{subject-relation} pair (Section \ref{sec:task_resource}).
    \item We experiment with a diverse set of \llm. Our experiments and analyses highlight how \llm\ poorly perform on \task\ (Section \ref{Experiments and Results:exp_result}).
    \item We propose \newmodel, a framework to improve temporally consistent factuality in \llm\ (Section \ref{sec:new_model}). Our experiments highlight that \newmodel\ surpasses the recent baseline models (Section \ref{Experiments and Results:exp_result}). We further analyze how the probabilistic space evolves under \newmodel\ (Appendix \ref{appendix:prob_abl}).
\end{itemize}

\begin{figure*}[!t]
\centering
\includegraphics[width=0.9\textwidth]{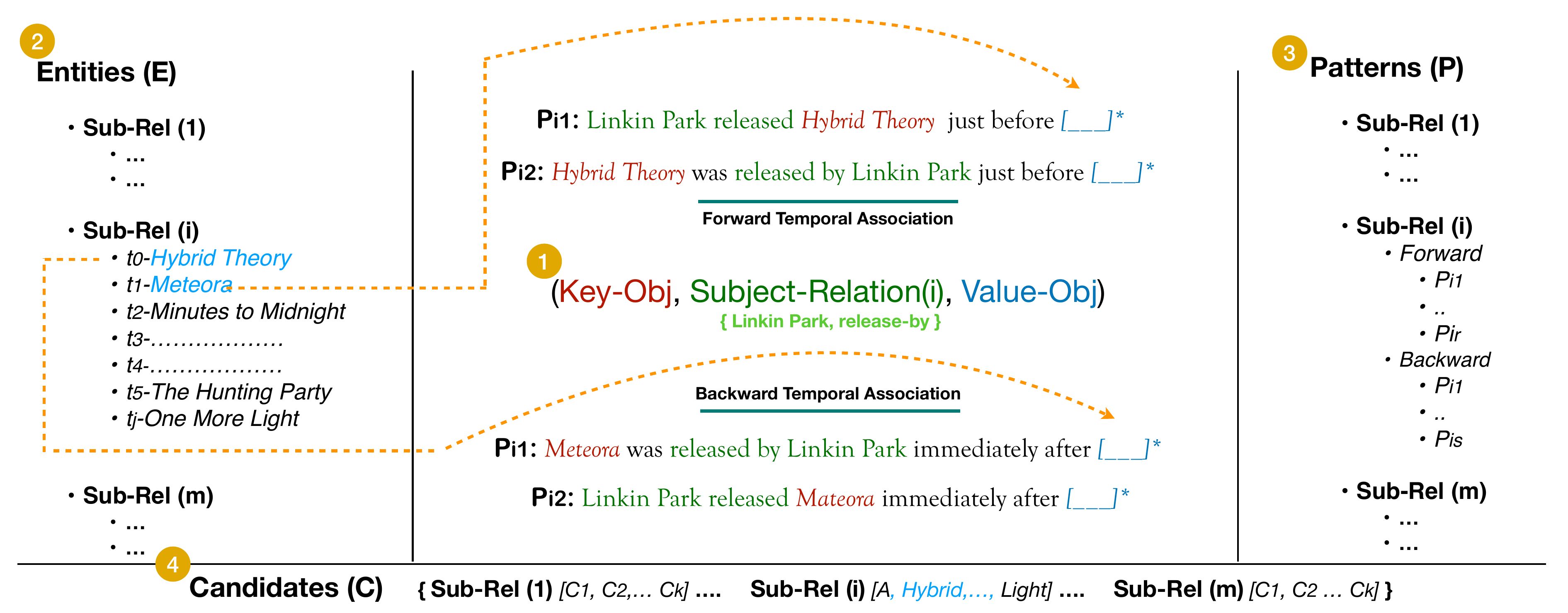}
\caption{The architectural framework of \newdataset\ -- (1) a set of diverse \textit{subject-relation} pairs, (2) a sequence of entities which are temporally connected via a given \textit{subject-relation} pair, (3) a set of paraphrase templates with a placeholder for \textit{key_object} and \textit{value_object} developed from \textit{subject-relation} pairs, and (4) a closed vocabulary candidate set developed from possible entity space for a given \textit{subject-relation} pair.}
\label{fig:taskarchitecture}
\end{figure*}

\section{The \newdataset\ Dataset}
\label{sec:task_resource}
Here, we present a novel English prefix-style \newdataset\ dataset with a temporal range of $1526$-$2022$. Inspired by \citet{elsahar-etal-2018-rex}, we semi-automatically\footnote{We leverage \gptfour\ to extend manually constructed initial set of base subject-relation pairs.} curate a diverse set of base subject and relations pairs \textit{subject-relation}. Next, we define $E_i$, a strict temporally ordered set of entities associated with $i^{th}$ \textit{subject-relation} pair. An entity can act as a \textit{key_object} or a \textit{value_object} relative to the role of another entity positioned right next/before it. We then create \textit{base\_patterns} for both forward and backward directions following \citet{petroni2019language}. Afterwards, a set of patterns $P_i$ is constructed by employing paraphrasing techniques \cite{bhagat-hovy-2013-squibs} on \textit{base\_patterns} in forward and backward directions followed by a candidate set $C_i$.

{\textbf{Construction approach.}} The \newdataset\ resource is constructed by three NLP experts\footnote{All of them are male with their ages ranging between 25-35 years.} with a mean cross-annotator agreement of $4.84 \pm 0.39$ and $4.86 \pm 0.49$ out of $5$ maximum on Likert scale \cite{likert} for factuality and consistency, respectively (refer Appendix \ref{appendix:data_abl} for more details). Following \citet{elazar2021measuring}, our construction process broadly follows a four-step procedure described in Figure \ref{fig:taskarchitecture}.

(i) First, we define a set of $m$ diverse \textit{subject-relation} pairs randomly collected from varied domains -- entertainment, technology, politics, automobiles and corporate. Using annotators' linguistic expertise, we then define \textit{base_patterns} for forward and backward directions, i.e.,
\textit{Linkin Park released [X] just before [Y]} 
and,
\textit{Linkin Park released [X] just after [Y]} 
are two \textit{base_patterns}  examples of the forward and backward temporal associations, respectively, where [X] and [Y] are the placeholders for a \textit{key_object} and a \textit{value_object}, respectively. 

(ii) Next, we manually curate a set of entities $E_i$ ($\forall i=0,\ldots,m$) for the $i^{th}$ \textit{subject-relation} pair such that they have temporal association with it in ascending temporal sequence $t=0,\ldots,j$, where, entity $e_i^t$ precedes $e_i^{t-1}$ and is followed by $e_i^{t+1}$ in the temporal space. The cardinality of $E_i$, represented by $j$, varies for each \textit{subject-relation} pair.

(iii) A set of paraphrased patterns $P_i$ ($\forall i=0,1,\ldots,m$) is constructed through an application of an online paraphraser tool, Quillbot\footnote{\url{https://quillbot.com/}} on \textit{base_patterns}. Set $P_i$ defines $r$ and $s$ number of paraphrases in the forward and backward directions, respectively. For simplicity, here we consider uniform values of $r$ and $s$ to be $8$. 

(iv) Finally, a constrained candidate set $C_i$ is developed as an unordered set of words from $E_i$. 

Table \ref{tab:data_desc} summarizes the statistics of \newdataset. Readers can refer to Appendix \ref{appendix:data_abl} for more detail about \newdataset, including annotation quality, temporal and entity type distributions.

\section{\task\ Task Structure}
\label{sec:task_framework}

 We pose \task\  as a sentence completion task that aligns with the general objective of an LM. With prefix style, paraphrase templates for a \textit{subject_relation} pair are filled with only \textit{key_object}, and we expect the model to generate \textit{value_object}. Next, we define \task\ metrics to evaluate \llm.

{\textbf{Temporal factuality and temporal consistency.}} We extend the metrics defined by \citet{elazar2021measuring} to \task. The first metric, \textit{temporal-factuality}, captures the accuracy of a model across the temporal direction. In contrast with their definition of exact-match accuracy, we define factuality as soft accuracy, a ratio of the number of continuous matches of words in actual and generated \textit{value_object} for a sample. It helps capture the partially correct generation as well. Next, we define its sub-classification across temporal directions. The forward \textit{temporal-factuality} measures the accuracy in the forward temporal direction where \textit{value_object} is located at $t+1$ time step for a given \textit{key_object} at $t$. In backward \textit{temporal-factuality}, the \textit{value_object} is located at $t-1$ time step for a given \textit{key_object} at $t$. 

\begin{table}[t!]
\small
\centering
\begin{tabular}{@{\extracolsep{10pt}}l|r}
\hline
\# Subject-Relation pairs	& 66\\
\hline
\# Paraphrase patterns & 	1056\\
  \hspace{5mm} \# Forward patterns	&528\\
  \hspace{5mm}  \# Backward patterns	&528\\
\hline
Avg \# pattern per relation-subject	&18\\
\hline
\# Entities	& 700\\
    \hspace{5mm}     \#Unique entity types 	& 11\\
    \hspace{5mm}     Min \# entities per relation-subject	& 2\\
    
    \hspace{5mm}     Max \# Entities per relation-subject	& 16\\
    
    \hspace{5mm}     Avg \# entities per relation-subject	& 10.6\\
\hline
\# Unique samples in dataset & 10144\\
\hline 
\end{tabular}
\caption{High-level statistics of the \newdataset\ dataset.}
\label{tab:data_desc}
\vspace{-3mm}
\end{table}

The second metric is \textit{temporal-consistency}. Given a pair of prefix-style paraphrases for a \textit{subject_relation} filled with an identical \textit{key_object}, an identical \textit{value_object} should be generated. The metric estimate is binary (one or zero) for a given pair of such paraphrases if the model's responses are identical or different. Forward and backward directions paraphrases contribute to forward and backward \textit{temporal-consistency}, respectively. 

{\textbf{Temporally consistent factuality.}} The third composite metric, \textit{temporally-consistent-factuality}, is a stricter version of \textit{temporal-factuality}, requiring a model to be consistent and factual across the temporal direction. It reports the \textit{temporal-factuality} only if the responses are identical from all prefix-style paraphrases for a given \textit{subject_relation} and  \textit{key_object} in a particular temporal direction. Paraphrases in forward and backward directions contribute to forward and backward \textit{temporally-consistent-factuality}, respectively.

{\textbf{Other metrics.}} The quality of patterns is measured using \textit{temporal-succ_patt}, indicating the percentage of patterns yielding a correct \textit{value_object} at least once during the probe. Furthermore, \textit{temporal-succ_objs} is introduced to measure the model's temporal knowledge by reporting the percentage of \textit{value_objects} accurately generated at least once. Further, we define \textit{temporal-know_cons} and \textit{temporal-unk_cons} as metrics to measure \textit{temporal-consistency} of the fraction of patterns which generated correct \textit{value_object} at least once and the fraction of patterns which never responded with a correct \textit{value_object} for a \textit{subject_relation}, respectively. All four metrics are then classified into forward and backward temporal directions.

\section{\newmodelfull\ (\newmodel)}
\label{sec:new_model}
Equipped with the advancements in model fine-tuning combined with the recent baseline in temporal reasoning, We begin with a base pre-trained LLM and apply supervised multi-task instruction-tuning to develop a multi-task instruction-tuned (MT-IT) model. Subsequently, we apply time and consistency-sensitive reinforcement learning to the MT-IT model to enhance its temporally consistent factuality capabilities.

{\textbf{Motivation.}} Instruction tuning (IT) \cite{zhang2023instruction} is a cost-effective, efficient technique for developing specialized models, as evidenced by the parameter-efficient fine-tuning (PEFT) \cite{peft} approach such as low-rank adaptation (LoRA) \cite{hu2021lora}, which offers significant benefits for instruction tuning in LLMs within low-cost infrastructures. Additionally, the combination of RL and supervised fine-tuning enhances model performance in several domain-specific tasks, notably temporal reasoning \cite{tan-etal-2023-towards}. Furthermore, advances in multi-task learning led to several breakthroughs in modeling multiple objectives simultaneously \cite{9392366}. 

\begin{figure}[!t]
\centering
\includegraphics[width=0.43\textwidth]{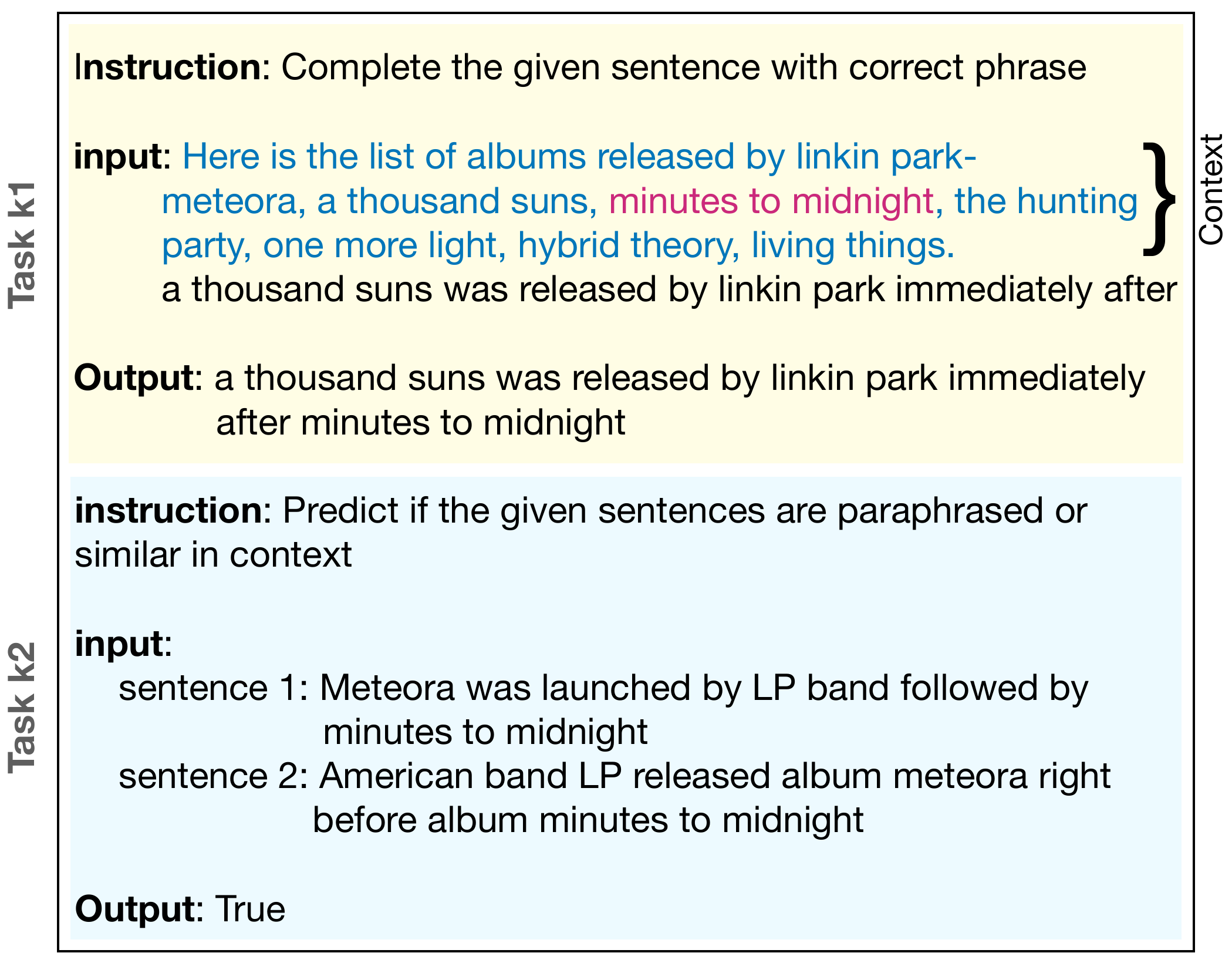}
\caption{An instruction-based sample from training data for MT-IT model. Task k1: Generative sentence completion; Task k2: Binary paraphrase prediction.}
\label{fig:mtit}
\vspace{-5mm}
\end{figure}

{\textbf{Multi-Task Instruction-Tuning (MT-IT).}} We consider a multi-objective optimization problem that simultaneously improves model's factuality and consistency, denoted by tasks k1 and k2, respectively. In k1, we apply a standard sentence completion task where an incomplete sentence filled with \textit{key_object}, augmented with a context and task instruction, is passed as input to allow the model complete the sentence with expected \textit{value_object}. Whereas k2 is a binary task predicting true or false if two sentences are paraphrased. We start with transforming the \newdataset\ dataset into an instruction-based dataset in line with the formats proposed by \citet{wang2023selfinstruct,alpaca} (Figure \ref{fig:mtit}). Next, we consider a base pre-trained LLM with parameter $\theta$ and then apply LoRA instruction-based supervised fine-tuning where the objective is to maximize ${p(o^{k1} | s^{k1}, i^{k1}, c)}$ for improving model's factual capability, where $s^{k1}$, $i^{k1}$, $c$ and, $o^{k1}$ are instruction, input, context and output, respectively for task k1. We further add another objective of maximizing ${p(o^{k2} | s^{k2}, i^{k2})}$ to improve the model's consistency in extending the framework to multi-task learning setup (Figure \ref{fig:mtit}), where $i^{k2}$ is a pair of sentences randomly sampled positively and negatively to predict the output $o^{k2}$ as true or false, respectively.

{\textbf{Consistent Time-Sensitive Reinforcement Learning (\ctsrl).}} Multiple \textit{value_object}s for a given \textit{key_object} are possible by excluding the temporal direction. We introduce \ctsrl, aimed at modeling consistent sensitivity towards time and overcoming the limitations of the binary temporal characterization inherent in TSRL  \cite{tan-etal-2023-towards}. \ctsrl\ is employed to further fine-tune $\theta$ through the joint modeling of both k1 and k2. The reward mechanism is devised as a linear amalgamation of temporal and consistence sensitivities aspects. Initially, $\ctsrl_{Discrete}$ is conceptualized in alignment with TSRL’s binary reward framework, awarding a positive discrete reward of one for accurate predictions and zero for incorrect responses. Additionally, $\alpha$ is defined to act as the weighting parameter for the consistence sensitivity reward component.
\begin{table*}[!t]
\small
\centering
\begin{tabular}{@{\extracolsep{10pt}}lccccccccc}
\hline

\multirow{3}{*}{\textbf{Models}}
      & \multicolumn{3}{c}{\textbf{Temp-fact}} &
      \multicolumn{3}{c}{\textbf{Temp-cons}} &
      \multicolumn{3}{c}{\textbf{Temp-cons-fact}}\\
       \cline{2-4}\cline{5-7}\cline{8-10}

& {\textbf{Avg}} &  \textbf{Bwd} & \textbf{Fwd} &  \textbf{Avg} &  \textbf{Bwd} & \textbf{Fwd} &  \textbf{Avg} &  \textbf{Bwd} & \textbf{Fwd} \\
\hline
GPT-J [6B] & 3.63&	6.88&	0.37&	64.87&	66.29&	63.46&	1.48&	2.73&	0.22\\
Falcon [7B] &2.99&	5.74&	0.24&	41.52&	40.42&	42.63&	1.08&	2.03&	0.13\\
LLaMA [7B] &1.48&	0.89&	2.07&	13.28&	12.52&	14.03&	0&	0&	0\\
LLaMA [13B] &1.11&	1.01&	1.21&	15.65&	16.33&	14.96&	0.2&	0.17&	0.24\\
LLaMA2 [7B] &0.95&	0.73&	1.17&	22.34&	21.95&	22.73&	0.08&	0&	0.17\\
LLaMA2 [13B] &1.13&	0.97&	1.27&	13.34&	14.41&	12.26&	0.09&	0&	0.19\\
   \hline
\end{tabular}

\caption{\label{tab:def_res_llm}
Zero-shot performance in open vocabulary setting on \task\ across various \llm. \textit{Temp-fact}: temporal factuality (in \%), \textit{Temp-cons}: temporal consistency (in \%), \textit{Temp-cons-fact}: temporally consistent factuality (in \%). \textit{Fwd} and \textit{Bwd} are the forward and backward direction probe, respectively (\textit{Avg}, an average of both directions).
}
\vspace{-3mm}
\end{table*}
\begin{equation}
\label{eq:rew_discrete}
R_{d}(x) = (1-\alpha) R_d^t(x) + \alpha R_d^c(x)
\end{equation}
\begin{equation}
\label{eq:rew_discrete_ind}
    R^\lambda_d(x) = \begin{dcases}
       P^\lambda_d(x), & \textit{if } {O_g}(\theta(x)) = O_l(x),\\
        N^\lambda_d(x), & otherwise. 
        \end{dcases}
\end{equation}
For a given input $x$, the overall reward function $R_d(x)$ in the $\ctsrl_{Discrete}$ setting is presented in Equation \ref{eq:rew_discrete}, where  $R_d^t(x)$ and $R_d^c(x)$ are reward contribution for tasks k1 and k2, respectively. In Equation \ref{eq:rew_discrete_ind}, $\lambda$ is used as a task indicator where $P^\lambda_d(x)$ is a positive reward score for consistence sensitivity component if $\lambda$ equals to $c$. Similarly, $N^\lambda_d(x)$ is a negative reward score. ${O_g}(\theta(x))$ and $O_l(x)$ are the generated \textit{value_object} and ground-truth label \textit{value_object}, respectively for given input $x$. In the case of $\ctsrl_{Discrete}$, a positive reward score equal to one is assigned in case of correctly generated output and zero otherwise for both tasks k1 and k2.

We further define another variant, $\ctsrl_{Smooth}$, to model continuous and relative properties of time. The temporal sensitivity reward is a continuous function where a positive reward has a maximum function value equal to one. However, the negative reward is proportional to the relative distance of the predicted answer from the correct answer in the temporal axis. The goal is to penalize incorrect answers that are distant from the correct answer more severely than the incorrect predictions that are nearby on the temporal axis. There is no change in the reward component for consistence sensitivity except for releasing the constraint of the parameter $alpha$, indicating that $\ctsrl_{Smooth}$ is an unweighted linear combination of continuous temporal and discrete consistence sensitivity reward components.
\begin{equation}
\label{eq:rew_smooth}
R_{s}(x) =  R_s^t(x) + R_s^c(x)
\end{equation}
\begin{equation}
\label{eq:rew_smooth_neg}
    N^t_s(x) = \begin{dcases}
    \frac{|t_{O_l} - t_{O_g}|}{t_n - t_{O_l}}, & \textit{if } t_{O_g} > t_{O_l}, \\
     \frac{|t_{O_l} - t_{O_g}|}{t_{O_l}}, & \text{otherwise} 
        \end{dcases}
\end{equation}
Equation \ref{eq:rew_smooth} presents reward function $R_s(x)$ in $\ctsrl_{Smooth}$ for a given input $x$, where $R_s^t(x)$ and $R_s^c(x)$ are the reward contribution for tasks k1 and k2, respectively. For the temporal sensitivity task (k1), the positive reward score is assigned as a value equal to one in case of correctly generated output, and the negative reward score is the relative distance of the wrong prediction from the correct prediction in the temporal axis as presented in Equation \ref{eq:rew_smooth_neg}. The symbol $t_{O_l}$ is the time step of the ground-truth label, $t_{O_g}$ represents the time step for generated output, and $t_n$ is the end time step of entity sequence for that \textit{subject-relation} pair.

\begin{figure}[!t]
\centering
\includegraphics[width=0.4\textwidth]{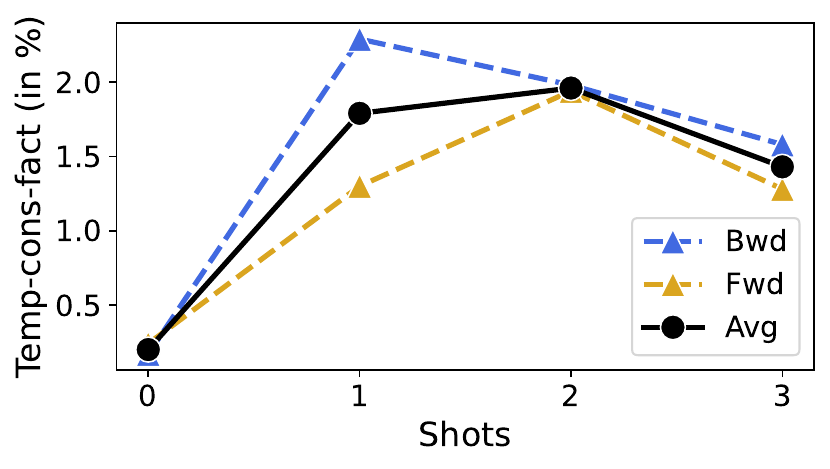}
\caption{Results for temporally consistent factuality (\textit{Temp-cons-fact}) in $k$-shot ($k$=1,2,3) ICL setup with \llama[13B] in an open vocabulary setting.}
\label{fig:shots_def_llama13}
\vspace{-5mm}
\end{figure}

\section{Experimental Results} 
\label{Experiments and Results:exp_result}

{\textbf{Experimental setup.}} The families of \gptj\footnote{\url{https://huggingface.co/EleutherAI/gpt-j-6b}}, \falcon\ \cite{almazrouei2023falcon}, and \llama\ \cite{touvron2023llama} are considered for evaluation on \task. Primarily, we evaluate \llm\ in an open vocabulary setting where the next token is sampled from the entire vocabulary. Next, we conduct experiments in an in-context learning setup, followed by a closed vocabulary setting. Finally, the \newmodel\ efficacy evaluation is conducted. 

{\textbf{Temporally consistent factuality on \task.}} We start with a comparison of various \llm\ in zero-shot setting. The task is to correctly complete a sentence with the expected \textit{value_object} given an instruction followed by an input, i.e., \textit{"complete the given sentence with the correct phrase: Meteora was released by Linkin Park immediately after"}. We observe the destitute performance of all the experimented \llm\ over both \textit{temporal-factuality} and \textit{temporally-consitent-factuality} in the range of [$0.95\%-3.63\%$] and [$0\%-1.48\%$], respectively (Table \ref{tab:def_res_llm}). We notice that \gptj\ and \falcon\ tend to be highly consistent in the range of [$41.52\%-64.87\%$] while being miserably factually incorrect compared to the \llama\ model in the same range of parameters size. Additionally, various families of \llm\ behave differently regarding their sensitivity towards temporal direction, but it doesn't significantly correlate with temporal consistency. Further, a preliminary evaluation of \task\ in zero-shot setting for commercial \llm\ such as \gptfour\ \cite{openai2023gpt4} and \claudethree\ is \cite{anthropic2024claude} presented in Appendix \ref{appendix:comm_llm}.

\begin{figure}[!t]
\centering
\includegraphics[width=0.45\textwidth]{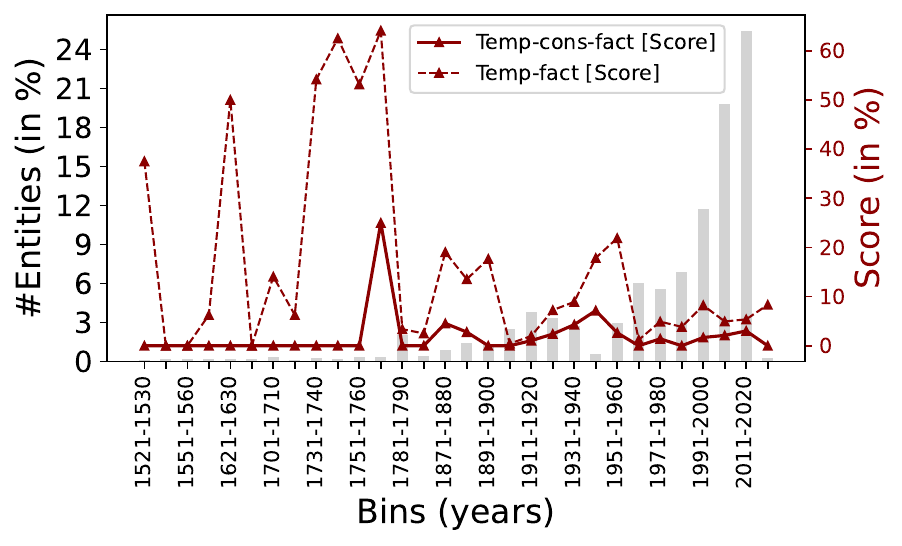}
\caption{Average temporally consistent factuality and temporal factuality (second y-axis) in an open vocabulary and two-shot setting across temporal bins of Entities (bin size: $10$ years) with \llama[13B].}
\label{fig:temp_abl}
\vspace{-6mm}
\end{figure}

{\textbf{In-context setup.}} In-context Learning (ICL) helps off-the-shelf \llm\ solve unseen tasks without the requirement of fine-tuning  \cite{dong2023survey}. We provide $k$ randomly-drawn examples from the same \textit{subject_relation} pair as supplementary context in a $k$-shot ICL setting, i.e., a one-shot example is as follows:  \textit{"complete the given sentence with the correct phrase: Meteora was released by Linkin Park immediately after => Hybrid Theory. American band LP released Minutes to Midnight immediately after =>"}. This test evaluates \llamathirteen\ and varies $k$ in the range [1-3]. In a two-shot setup, we observe absolute percentage points improvement of $1.76$ in \textit{temporally-consistent-factuality} (Figure \ref{fig:shots_def_llama13}). At the same time, the improvements of $7.01$ and $18.22$ percentage points are noted in contributory metrics \textit{temporal-factuality} and \textit{temporal-consistency}, respectively (refer to Appendix \ref{appendix:icl_abl} for more details).

Figure \ref{fig:temp_abl} presents \textit{temporal-factuality} across the temporal distribution of entities. Findings reveal that \textit{temporal-factuality} for entities belonging to the historical period ($1500$-$1800$) is significantly higher, with an average of $25.27\%$ compared to $9.08\%$ for the entities in the contemporary period (after $1800$). At the same time, the presence of multiple sources of the same information in \llm\ pre-trained dataset for the contemporary period leads to better \textit{temporally-consistent-factuality}.

\begin{figure}[!t]
\centering
\includegraphics[width=0.45\textwidth]{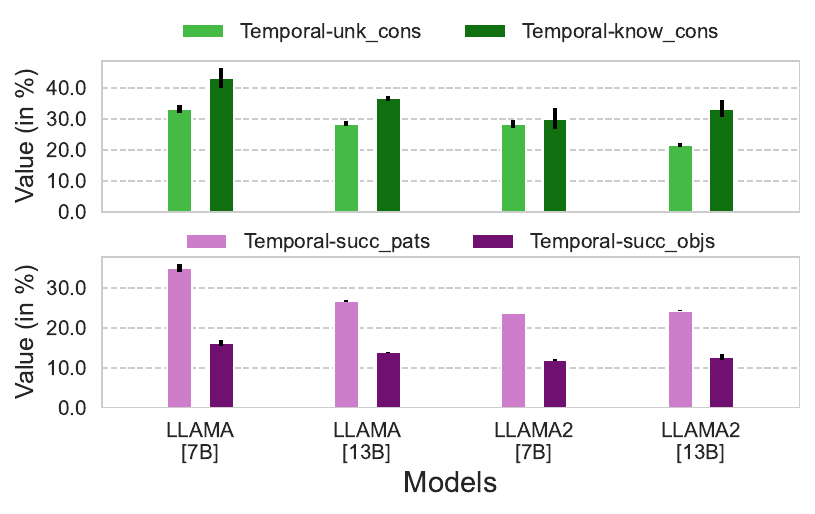}
\caption{Qualitative results in an open vocabulary with two-shot ICL setup across \llama\ variants. Error bars represent divergence across temporal directions -- forward and backward.}
\label{fig:qualityAna}
\vspace{-5mm}
\end{figure}

Additionally, in Figure \ref{fig:qualityAna}, qualitative analysis reveals that \llamaseven\ attains best \textit{temporal-know_cons} at $43.29\%$ with a divergence of $10.00$ percentage points between known and unknown temporal consistencies across \llama\ variants. On the other hand, $35.04\%$ patterns yield a correct \textit{value_object} at least once in contrast to only $16.18\%$ \textit{value_objects}, which were predicted correctly once in the entire probe for a model.

\begin{figure}[!t]
\centering
\includegraphics[width=0.43\textwidth]{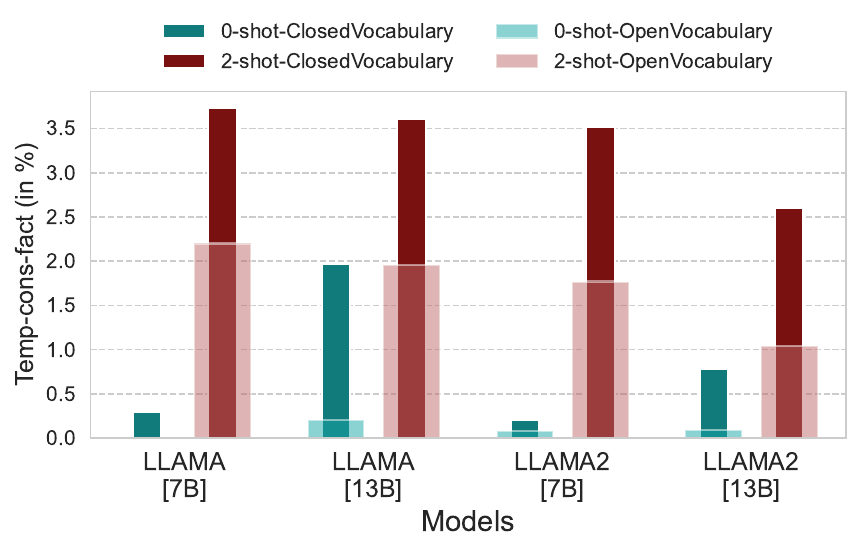}
\caption{A comparison of average \textit{temporally-conistent-factuality} between open and closed vocabulary settings across \llama\ variants in zero-shot and 2-shot
.}
\label{fig:openCloseVocab}
\vspace{-5mm}
\end{figure}

{\textbf{Closed vocabulary setup.}} The next word generated by an LM can still be the right placement given its general objective to maximize the semantic expectation irrespective of the expected \textit{value_object}. Therefore, we also conduct experiments in a closed vocabulary setting where the sample space is reduced to a candidate set\footnote{A set of restricted tokens are generated by employing byte pair encoding \cite{sennrich2016neural} on candidate set.} (defined in Section \ref{sec:task_resource}) during generation. This approximation helps set up the probe as KB fact extraction from a given possible facts space, thus maximizing the behavioral expectation of LM as KBs. With an improvement in the range [$1.77\%$ - $2.08\%$], \llamathirteen\ has best scores of $1.97\%$ and $3.51\%$ for \textit{temporally-consistent-factuality} in one and two shots setting, respectively, across variants of \llama\ under closed vocabulary setting (Figure \ref{fig:openCloseVocab}). We observe a similar trend for \textit{temporal-factuality} and \textit{temporal-consistency}, presented in Appendix \ref{appendix:closed_abl}.

{\textbf{Improvements with \newmodel.}} We conduct all experiments in an open vocabulary setting (assuming no access to candidate sets during inferences) with \llamathirteen. First, \newdataset\ is vertically split (test ratio: $0.3$) to produce a train set and a test set containing 46 and 26 subject-relation pairs, respectively. The random vertical split ensures the stricter evaluation of \newmodel\ as the test set contains only unseen \textit{subject-relation} pairs. We broadly categorize this evaluation into four categories: (i) the default performance of the model in zero-shot and two-shot (ICL) setup, (ii) variants of instruction-tuned (IT) models based on the presence of a context along with the novel multi-task IT model, (iii) IT (with context) followed by TSRL; a strong baseline model, and (iv) \newmodel, a combined strategy of MT-IT followed by variants of novel \ctsrl\ method. Empirically, we set $\alpha$ as $0.66$ in formulating a discrete variant of \ctsrl\ (significance of $\alpha$ is presented in Appendix \ref{appendix:alpha_abl}).

\begin{table*}[!ht]
\small
\centering

\begin{tabular}{@{\extracolsep{1pt}}l@{\hspace{0.3cm}}l@{\hspace{0.4cm}}l@{\hspace{0.2cm}}l@{\hspace{0.2cm}}ll@{\hspace{0.2cm}}l@{\hspace{0.2cm}}ll@{\hspace{0.2cm}}l@{\hspace{0.2cm}}l}
\hline

\multirow{3}{*}{\textbf{Models}} & \multirow{3}{*}{\textbf{Setting (open vocab)}}
      & \multicolumn{3}{c}{\textbf{Temp-fact}} &
      \multicolumn{3}{c}{\textbf{Temp-cons}} &
      \multicolumn{3}{c}{\textbf{Temp-cons-fact}}\\
       \cline{3-5} \cline{6-8} \cline{9-11}

& & \textbf{Avg} &  \textbf{Bwd} & \textbf{Fwd} &  \textbf{Avg} &  \textbf{Bwd} & \textbf{Fwd} &  \textbf{Avg} &  \textbf{Bwd} & \textbf{Fwd} \\

\hline

\multirow{2}{*}{Default} & Zero-shot &${0.40}_{\pm0.00}$ &0.28	&0.52&	${16.31}_{\pm0.00}$	&15.22&	17.40&	${0.00}_{\pm0.00}$	&0.00&	0.00  \\
 & Two-shot [ICL] &${4.95}_{\pm0.40}$	&4.26	&5.63	&${31.57}_{\pm1.50}$	&29.23	&33.91	&${0.53}_{\pm0.18}$	&0.50	&0.57 \\
   \hline
\multirow{2}{*}{IT} & Without context 
&${11.53}_{\pm0.08}$&	10.36&	12.69&	${25.08}_{\pm0.06}$&	26.76&	23.41&	${2.59}_{\pm0.30}$&	2.67&	2.50\\
 & + Context &${16.87}_{\pm0.17}$&	16.79&	16.96&	${34.39}_{\pm0.92}$&	34.45&	34.31&	${1.75}_{\pm0.15}$&	2.11&	1.38  \\
  \hline
MT-IT & + Context &${18.17}_{\pm0.15}$&	17.83&	18.52&	${36.60}_{\pm0.36}$&	36.49&	36.70&	${3.33}_{\pm0.41}$&	2.82&	3.85 \\
 \hline
 
$\Delta$a & &\textcolor{blue}{$1.3\uparrow$} & & &\textcolor{blue}{$2.21\uparrow$} & & &\textcolor{blue}{$1.58\uparrow$} & &\\

   \hline
\multirow{1}{*}{Baseline} & IT (+context) + TSRL  &${16.60}_{\pm0.51}$	&17.04	&16.15	&${33.23}_{\pm0.38}$	&33.28	&33.18	&${2.28}_{\pm0.44}$	&2.88	&1.67\\
\hline
   \multirow{2}{*}{\newmodel} & MT-IT+$\ctsrl_{Discrete}$ &$\textbf{18.72}^{\textcolor{teal}{P_1}}_{\pm0.25}$	&19.16	&18.29	&$\textbf{36.88}^{\textcolor{teal}{P_2}}_{\pm0.49}$	&37.58	&36.17	&$\textbf{4.34}^{\textcolor{teal}{P_3}}_{\pm0.27}$	&4.45	&4.22\\
 & MT-IT+$\ctsrl_{Smooth}$ &${18.16}_{\pm0.05}$	&18.11	&18.21	&${36.07}_{\pm0.48}$	&37.09	&35.04	&${3.89}_{\pm0.92}$	&3.74	&4.05\\
 \hline
 
$\Delta$b  & &\textcolor{blue}{$2.12\uparrow$} & & &\textcolor{blue}{$3.65\uparrow$} & & &\textcolor{blue}{$2.06\uparrow$} & &\\
\hline
&&&&&&\multicolumn{5}{r}{$\textcolor{teal}{P_1=0.003, P_2=0.003, P_3=0.003}$}
\end{tabular}
\vspace{-3mm}
\caption{\label{tab:main_res_new_model}
Experimental results of \newmodel\ across -- \textit{temporal-factuality}, \textit{temporal-consistency}, and \textit{temporally-consistent-factuality} (in \%), in comparison to multiple baselines on test data with \llama[13B] (average scores over three runs). $\Delta$a: improvements of MT-IT over an IT model, $\Delta$b: improvements of \newmodel\ over a baseline model. \textcolor{teal}{($P_1, P_2, P_3$)}: $p$-values at \newmodel's best scores compared to baseline model with $n=3$ and one-tailed test.
}
\vspace{-3mm}
\end{table*}

In Table \ref{tab:main_res_new_model}, the additional context provided with an input improves \textit{temporal-factuality} by $5.22$ percentage points for an IT model. The MT-IT model improves \textit{temporal-factuality}, \textit{temporal-consistency} and \textit{temporally-consitent-factuality} by $7.7\%$, $6.4\%$ and $90.2\%$, respectively, compared to the IT model. We further observe improvements of $12.7\%$, $10.9\%$ and $90.4\%$ in \textit{temporal-factuality}, \textit{temporal-consistency} and \textit{temporally-consitent-factuality}, respectively, with \newmodel\ over the baseline, indicating that improving a model's temporal consistency also positively impacts its temporal factuality. However, the inaccessibility of \gptfour\ architecture limits us to assess the efficacy of \newmodel's on this model. (the probabilistic space evolution under \newmodel\ including the ablations for scalability are presented in Appendix \ref{appendix:prob_abl}, \& \ref{appendix:scalability}).

{\textbf{Significance of \ctsrl\ over TSRL.}} We perform this ablation in two different scenarios by immobilizing the SFT model. In the first scenario, an IT model serves as the foundation [comparing (IT + TSRL) with (IT + $\ctsrl_{Discrete}$)], while in the second scenario, the basis is the MT-IT model [comparing (MT-IT + TSRL) with (MT-IT + $\ctsrl_{Discrete}$)]. These experiments are meticulously executed under identical conditions, as meticulously outlined in Table \ref{tab:main_res_new_model}.

\begin{table}[!t]
\small
\centering
{
\begin{tabular}{lccc}
\hline
\textbf{Model} & \textbf{Temp-fact} &  \textbf{Temp-cons} &  \textbf{Temp-cons-fact} \\
\hline
\\
\multicolumn{3}{l}{Base SFT Model: IT}\\
\hline
TSRL&16.60&	33.23&	2.28 \\
\ctsrl&\textbf{17.6}&\textbf{34.27}&\textbf{2.74} \\
\hline
\\
\multicolumn{3}{l}{Base SFT Model: MT-IT}\\
\hline
TSRL&17.55&	34.5&3.68\\
\ctsrl&\textbf{18.72}&\textbf{36.88}&\textbf{4.34}\\
\hline
\end{tabular}}
\caption{\label{tab:ctsrl_tsrl}
Comparison of $\ctsrl_{Discrete}$ and TSRL by immobilizing the SFT model across two settings: IT and MT-IT model.}
\vspace{-3mm}
\end{table}

The findings are delineated in Table \ref{tab:ctsrl_tsrl}. $\ctsrl_{Discrete}$ exhibits superior performance over TSRL on all metrics: \textit{temporal-factuality}, \textit{temporal-consistency}, and \textit{temporally-consistent-factuality} in both experimental settings. Results from this ablation study further underscore the distinct advantage of a preference of \ctsrl\ over TSRL.

\section{Error Analysis}
\label{sec:err_ana}
\begin{figure}[!t]
\centering
\includegraphics[width=0.45\textwidth]{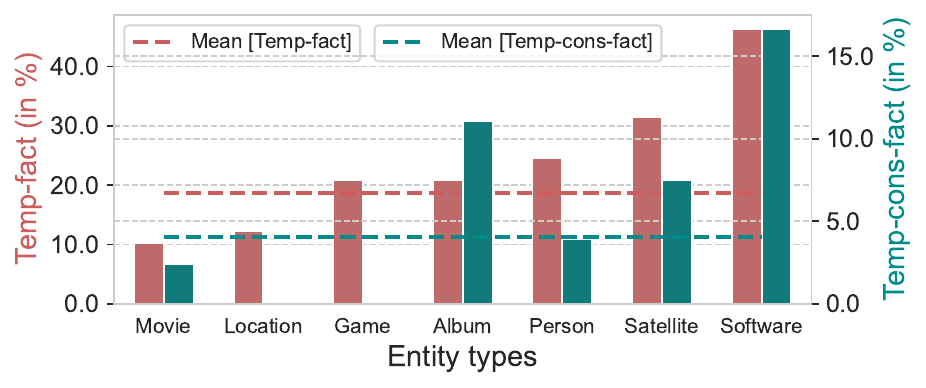}
\caption{Temporally consistent factuality across various entity-types present in test data for $\ctsrl_{Discrete}$.}
\label{fig:error_ana2}
\vspace{-5mm}
\end{figure}

A causal analysis is conducted to determine any correlation between data characteristics and failure cases. Figure \ref{fig:error_ana2} shows a strong correlation between entity-type and \textit{temporally-consistent-factuality}. Entity types not exclusively attached to a \textit{subject-relation} pair, such as movie names and geographical locations, perform poorly compared to other entities, such as satellite, person, and software names. It would be interesting to enhance both \newdataset\ and \ctsrl\ formulation to include such data characteristics explicitly. For more detail on this, readers can refer to Appendix \ref{appendix:error_abl}.

\section{Related Work}
\label{sec:rel_work}
A noteworthy advancement in temporal factuality involves analyzing and updating \llm\ to address the obsolescence of their factual knowledge over time \cite{hu2024towards,vu2023freshllms}. These approaches concentrate on amending the factuality of evolving temporal relationships without rectifying the current inaccuracies and inconsistencies. On the other hand, consistency in KBs was extensively studied, with developments around data and methods to benchmark the degree of inconsistencies \cite{Hansen2000,article,THIMM20131} across several tasks; QA \cite{kassner-etal-2020-pretrained}, reading comprehension \cite{7410636,rajpurkar-etal-2016-squad,ribeiro-etal-2019-red}, summarizing \cite{xie-etal-2021-factual-consistency, roit2023factually} and NLI \cite{li2019logicdriven}.

Notable works in temporal information extraction include TimeBank \cite{articlepus} and TimeEval \cite{verhagen-etal-2010-semeval}. The annotations defined in these datasets are primarily for time-event and events relationships such as before/after. The remarkable advancement in studying temporal reasoning over KGs led to the development of datasets such as TEQUILA \cite{10.1145/3269206.3269247}, TimeQuestions \cite{Jia_2021}, and CronQuesions \cite{saxena-etal-2021-question}, probing KG's response in ranking entities for a given temporal query. Subsequently, datasets like TEMPLAMA \cite{Dhingra_2022} and StreamingQA \cite{liška2022streamingqa} were curated for temporal reasoning in LMs. The TSQA \cite{chen2021dataset} dataset has addressed the drawback of the limited time span of queries, but it defines only time-event relationships. TEMP-REASON \cite{tan-etal-2023-towards} captures multiple facets of temporal reasoning -- time-time, time-event, and event-event relationships over a longer time span. Our dataset, \newdataset, addresses the lack of consistent temporal reasoning evaluation and improves upon TEMP-REASON by offering a diverse set of anchor queries through sixty-six unique sub-rel-obj triplets across eleven entity types, overcoming TEMP-REASON's limitation of single-entity-type anchors and coverage of only six entity types. (Appendix \ref{appendix:data_abl}).
 
\section{Conclusion}
\label{sec:conclusion}

This paper presented a new task \task\ with a novel resource, \newdataset, to evaluate temporally consistent factuality in large language models. The contribution continued to present a novel solution \newmodel\ based on multi-task instruction tuning (MT-IT) combined with consistent-time-sensitive reinforcement learning (\ctsrl) to improve \llm\ temporally consistent factuality. We observed that \newmodel\ outperforms the baselines to improve \textit{temporally-consistent-factuality} in \llm. The contribution and findings in this paper would help better understand and enhance \llm' underlying capabilities around the tasks which require consistent temporal reasoning and deductions. 

\section{Limitations}
\label{sec:limitations}
The scope of \newdataset\ is to evaluate \llm\ for their temporally consistent factuality capabilities. \newdataset\ comprehensively covers entity-entity temporal relations but falls short of stating entity-time and time-time aspects of temporal relations. Therefore, it should be applied along with other prominent datasets to estimate the overall temporal reasoning capabilities of \llm.

Moreover, the underlying rationale for the smooth variant of \ctsrl\ posits that a fundamental characteristic of time is its relativity. The objective is to impose greater penalties on incorrect responses that significantly deviate from the correct answer compared to those incorrect predictions that are proximal on the temporal scale. This premise suggests that such a formulation will compel the model to assimilate the relative aspect of time, thereby enhancing its efficiency in addressing queries necessitating relative temporal responses (before/after). While the smooth variant of \newmodel\ appears promising at the conceptual level, it has not surpassed the performance of its discrete counterparts. A comprehensive ablation study on smooth variants is earmarked for future investigation.

While temporal relation settings such as t+/-1 scenarios necessitate precision, as only a single correct response is viable for any given query contrary to the open temporal relation settings (one-to-many), it is our fervent hope that the insights derived from this study will inspire further research to test TeCFaP toward addressing multi-hop temporal queries (t+/-n) in forthcoming endeavors. On the other hand, the challenges related to computational assets to stack \llm\ continue. Due to asset limitations, we may not utilize more extensive or commercially accessible \llm\ to comprehensively evaluate on \task\ and approve on the off chance that the preferences of \newmodel\ are also advantageous to those models.

Furthermore, despite efforts to apply higher quality standards, \newdataset\ relies on human annotation and is therefore prone to annotation errors. Moreover, the base language of \newdataset\ is English; therefore, it falls short in measuring consistent temporal factuality for other languages, particularly low-resource ones. Extensions to multilingual setting or resource-poor languages are left to future research.

\section{Ethics Statement}
\label{sec:ethics}

The \newdataset\ is based on the Wikipedia and open world wide web knowledge sources. Wikipedia articles are licensed under a  Creative Commons Attribution-ShareAlike 4.0 International License\footnote{\url{https://en.wikipedia.org/wiki/Wikipedia:FAQ/Copyright}} (CC BY-SA 4.0) and its knowledge base is in the public domain. We will release \newdataset\ under same licence too. The experiments are conducted with all open source \llm. Authors do not intend to introduce biases in any form to \llm\ While applying fine-tuning methods.

The subjects and entities selected to be part of \newdataset\ are prone to unintended human biases during construction. The authors do not propagate any views/opinions, products, or representations of these subjects or entities in any form. The fact that women do not find representation during \newdataset\ annotations should be seen as a symptom of the gender disparity in research and innovation worldwide, but this is not the authors' view. We support gender and racial equality in research and innovation with the utmost sincerity. Furthermore, no generative AI-based content creation tools or applications were used to create this artifact, except for specialized support for spell checking, grammar correction, and paraphrasing. 

\section*{Acknowledgement}
Tanmoy Chakraborty acknowledges the support of the IBM-IITD AI Horizons network.

\bibliography{custom}
\bibliographystyle{acl_natbib}

\appendix
\section{Appendix}
\label{sec:appendix}

\subsection{Extended Description for \newdataset}
\label{appendix:data_abl}

In this section we continue to present details on \newdataset\ dataset. \newdataset\ is the first dataset of its kind in the temporal consistency domain, providing a strictly homogeneous sequence of entities for diverse subject-relation pairs. The conditioning on events that occurred without recurrence for strict temporal relationships along with the prefix formulation of {key-object, sub-rel, value-object} posed a severe challenge and required significant human evaluations at the level of \textit{subject-relation} pairs during construction.  

\begin{figure}[!h]
\centering
\includegraphics[width=0.95\columnwidth]{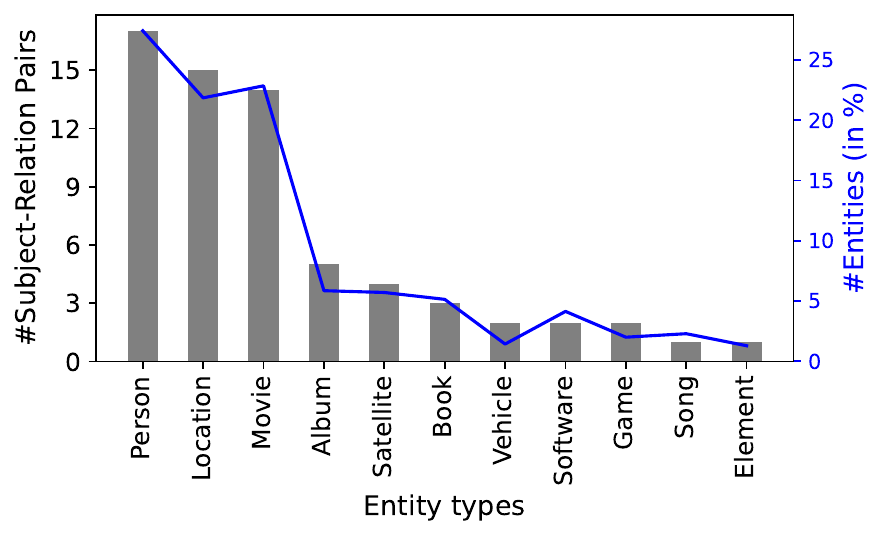}
\caption{Entity types distribution in \newdataset\ resource as \% of overall entities in data and representation across subject-relation (\textit{subject-relation}) pairs.}
\label{fig:ent_type}
\vspace{-5mm}
\end{figure}

{\textbf{Entity Types Distribution.}} We present the overall entity types distribution across different \textit{subject-relation} pairs in Figure \ref{fig:ent_type}. \newdataset\ has broader representations for various entity types such as person names, countries, books, movie and album names, satellites, vehicles, and software. 

{\textbf{Temporal Distribution.}} The temporal distribution of entities is presented in Table \ref{tab:temp_dist}. We observe that the entities have a more comprehensive range of temporal representations in \newdataset\ dataset spanning in the range of 1500 to 2022. Most of the pre-training datasets that \llm\ are trained on have a cutoff year of 2020, with few exceptions. Therefore, we consider that $99\%$ of all entities in \newdataset\ dataset must belong to a year equal to or less than 2020. It can be observed that there is a skewed temporal distribution of entities in favor of entities in the contemporary period (the year 1800-) compared to the historical period (the year 1500-1800). It is noted that $57\%$ of all entities are from 1991 to 2020.

\begin{table}[t!]
\small
\centering
\begin{tabular}{l|r|l|r}
\hline
\textbf{Bins (year)}&\textbf{\#Entities} &\textbf{Bins (year)}&	\textbf{\#Entities}\\
\hline
1521-1530	&	0.08	&	1881-1890	&	1.42\\
1531-1540	&	0.16	&	1891-1900	&	1.34\\
1551-1560	&	0.16	&	1901-1910	&	2.52\\
1601-1610	&	0.16	&	1911-1920	&	3.79\\
1621-1630	&	0.16	&	1921-1930	&	3.31\\
1651-1660	&	0.16	&	1931-1940	&	2.76\\
1701-1710	&	0.32	&	1941-1950	&	0.55\\
1711-1720	&	0.08	&	1951-1960	&	3.00\\
1731-1740	&	0.24	&	1961-1970	&	6.07\\
1741-1750	&	0.16	&	1971-1980	&	5.60\\
1751-1760	&	0.32	&	1981-1990	&	6.86\\
1761-1770	&	0.32	&	1991-2000	&	11.75\\
1781-1790	&	2.05	&	2001-2010	&	19.79\\
1791-1800	&	0.39	&	2011-2020	&	25.39\\
1871-1880	&	0.87	&	2021-2030	&	0.24\\
\hline 
\end{tabular}
\caption{Temporal distribution of entities (in \%) in \newdataset\ with a bin size of $10$ years.}
\label{tab:temp_dist}
\vspace{-3mm}
\end{table}

\begin{figure}[!t]
\centering
\includegraphics[width=0.9\columnwidth]{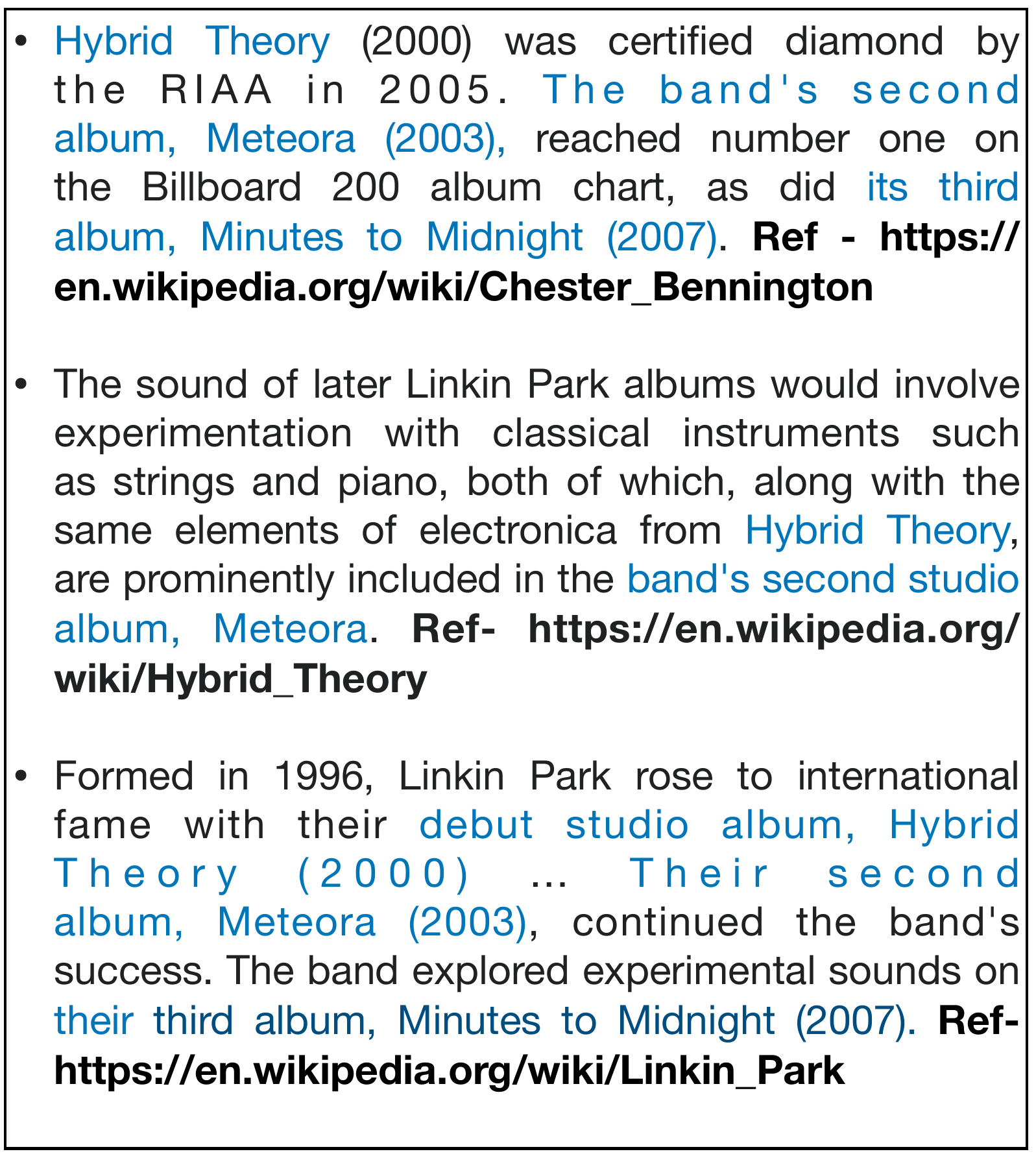}
\caption{A few extracts from Wikipedia for a \textit{subject-relation} pair (linkin park - release-by). We find the presence of sentences in pre-training data (Wikipedia) of \llm, which defines the temporal relationship among entities associated with the given \textit{subject-relation} pair.}
\label{fig:data_evi}
\vspace{-5mm}
\end{figure}

{\textbf{Evidences from Pre-training Data.}} \task\ is a novel task of evaluating the consistent temporal relationship between entities in \llm. Here, we present a few manually extracted evidence from pre-training data to support that the objective of \task\ is fairly expected from \llm. We consider Wikipedia for this test as it is a part of the pre-training dataset for most of the recent \llm, including \llama. Evidences are manually extracted from Wikipedia for an entity pair \textit{hybrid-theory and meteora} for a \textit{subject-relation} pair \textit{Linkin Park} and \textit{Release}. Figure \ref{fig:data_evi} presents the evidence with their sources in the pre-training dataset. Given these sentences, a human can easily find the temporal relation between  \textit{hybrid-theory} and \textit{meteora}. Therefore, it is a fair ask from \llm\ to learn the temporal relation between entities given such sentences.

{\textbf{Annotations Quality.}} We ensure the high quality of \newdataset\ resource by applying it to execute cross-annotator agreement experiment. To assess the qualitative measure of factuality, We randomly select a hundred samples of filled patterns across different \textit{subject-relation} pairs. The two reviewers split the data in half and reviewed the samples that the other annotator produced during construction. They score the agreements on a scale of 5-point (1-lowest agreement and 5-complete agreement on factuality of filled pattern) Likert scale \cite{likert}. Similarly, to give the qualitative measure for consistency of patterns created, we again randomly sample hundreds of filled patterns and pair these positively with another paraphrased filled pattern sampled randomly from respective \textit{subject-relation}. The reviewers repeat the similar scoring methodology in factuality assessment (1-lowest agreement and 5-complete agreement if the two patterns are paraphrased). We observe a mean agreement of $4.84 \pm 0.39$ out of $5$ maximum for factuality and a mean agreement of $4.86 \pm 0.49$ on similar lines for consistency.

{\textbf{\newdataset\ Coverage.}} We present the \newdataset\ comparison with prior datasets in Table \ref{tab:data_comp}. Some of the columns data reused from \citet{tan-etal-2023-towards} comparison of datasets. TEMPREASON is one of the most comprehensive temporal reasoning datasets, with 21K queries of event-event probe type in the QA setting. Here are the six types of entities it has considered during automatic construction from Wikipedia: Person, School, Political party, Company, Position, and Sports team. A significant drawback of TEMP-REASON is that all the queries are anchored around just person names, either as a subject or as an object. I.e., \textit{Which team did <subject> play for before/after oj?} or \textit{Who was the head of the government of <subject> before/after oj?}. 

In comparison, the proposed \newdataset\ has 10.1K prefix-style queries and covers eleven diverse entity types from several domains: Movies, Geographical location, Games, Albums, Persons, Satellites, Software, Books, Vehicles, Songs, and Elements. Here, the queries are anchored around eleven entity types via sixty-six diverse subject-relation pairs, such as \textit{Linkin Park released Hybrid Theory just before ___} or \textit{The FIFA U-17 World Cup was hosted by Canada immediately after ___}. 

The dataset will facilitate the further development around various aspects of consistent temporal reasoning such as consistent event sequencing, multi-hop sequence-based QA, consistent multi-reasoning QA, and architectural study of LLMs through temporal explainability of inconsistent behaviour via paraphrased queries.

\begin{table*}[!t]
\tiny
\centering
\begin{tabular}{lllllll}
\hline
\textbf{Dataset}& \textbf{Format}&  \textbf{Knowledge} &\textbf{Paraphrases} &\textbf{Temporal Sequence of} &\textbf{Time} &\textbf{Size} \\ 
&&\textbf{Source}&&\textbf{Homogeneously Grouped}&\textbf{Coverage}&\\
&&&&\textbf{Entities}&&\\
\hline
TEMPCOFAC& Prefix-Style:Language& Human Annotated/Open Web&\checkmark& \checkmark&1526-2022 &10.1K \\
TEMPREASON\cite{tan-etal-2023-towards}& QA:Language& Wikipedia/Wikipedia&\xmark& \xmark& 634-2023 &52.8K\\
TEMPLAMA\cite{Dhingra_2022}& QA:Language&Wikipedia& \xmark& \xmark&  2010-2020& 50k\\
Time-SensitiveQA\cite{chen2021dataset}& QA:Language&Wikipedia/Wikipedia& \xmark& \xmark& 1367-2018& 41.2k\\
StreamingQA\cite{liška2022streamingqa}& QA:Language&WMT& \xmark& \xmark&2007-2020&147k\\
\hline
TempQuestions\cite{Jia_2018}& QA:KG&Freebase& \xmark& \xmark &NA& 1.2k \\
TimeQuestions\cite{Jia_2021}& QA:KG&Wikipedia& \xmark& \xmark&  NA& 16.1k\\ CronQuestions\cite{saxena-etal-2021-question}& QA:KG&Wikipedia& \xmark& \xmark&34-2021& 410k\\
\hline
\end{tabular}
\caption{\label{tab:data_comp}
{Comparison of TEMPCOFAC with prior datasets.}
}
\vspace{-3mm}
\end{table*}

\subsection{Extended Results}

\subsubsection{ICL Setting Cont.}
\label{appendix:icl_abl}
\begin{figure}[!ht]
 \captionsetup[subfigure]{justification=centering}
    \centering
    \subfloat[Temporal factuality]{\includegraphics[width=0.22\textwidth]{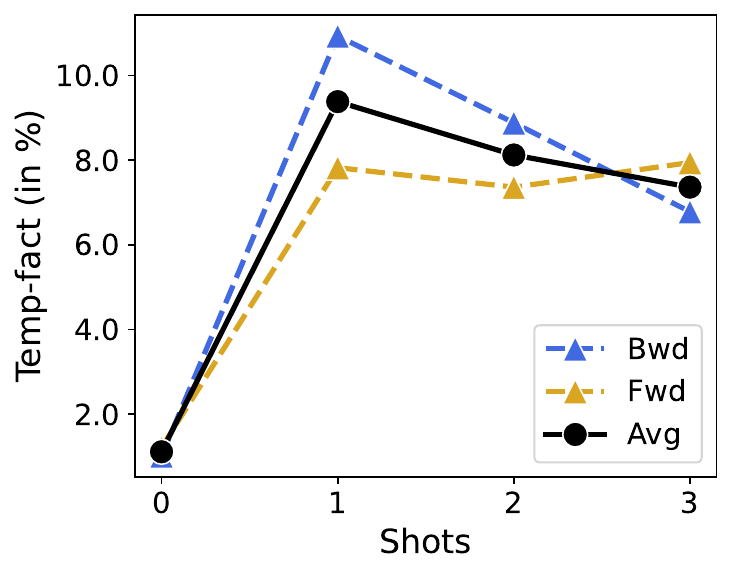}}\hspace{4mm} 
    \subfloat[Temporal consistency]{\includegraphics[width=0.22\textwidth]{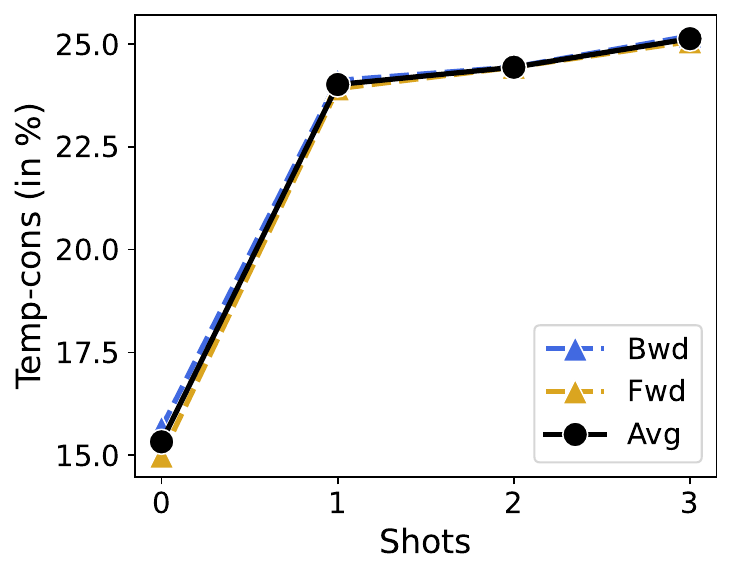}}\hspace{7mm} 
    \caption{Results for k-shot (k=1,2,3) ICL setup with \llama[13B] in an open vocabulary setup across -- (a) \textit{Temp-fact}: temporal factuality, (b) \textit{Temp-cons}: temporal consistency.}
    \label{fig:shots_def_llama13_abl}
\vspace{-5mm}
\end{figure}

Here, we continue from the result section and present results for \textit{temporal-factuality} and \textit{temporal-consistency} in ICT setting (Figure \ref{fig:shots_def_llama13_abl}).

\subsubsection{Closed Vocabulary Setting Cont.}
\label{appendix:closed_abl}

\begin{figure}[!h]
 \captionsetup[subfigure]{justification=centering}
    \centering
    \subfloat[Temporal Factuality]{\includegraphics[width=0.45\textwidth]{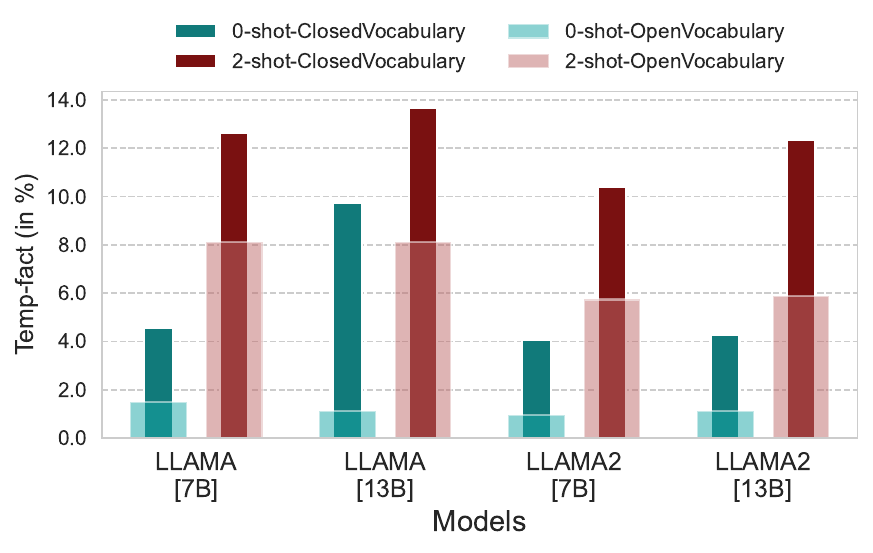}}\vspace{1mm} 
    \subfloat[Temporal Consistency]{\includegraphics[width=0.45\textwidth]{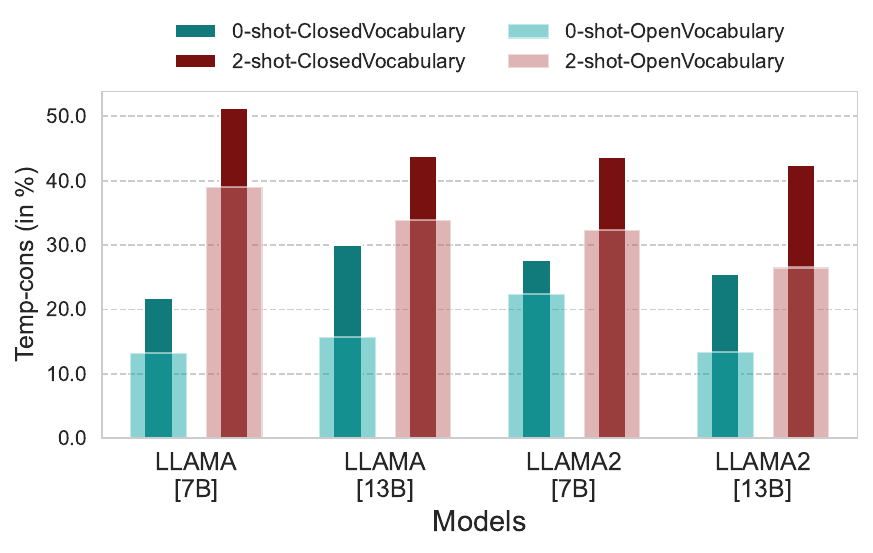}}  
    \caption{A comparison between open and closed (using restricted candidate set) vocabulary settings across variants of \llama\ model in zero-shot and in-context learning setup with k-shot (k=3).}
    \label{fig:openCloseVocabAbl}
\vspace{-5mm}
\end{figure}

We continue from the Section \ref{Experiments and Results:exp_result} to present the results for metrics \textit{temporal-factuality} and \textit{temporal-consistency} in closed vocabulary setting in Figure \ref{fig:openCloseVocabAbl}. Three-shot with closed vocab setting significantly outperforms by attaining maximum scores of $12.79\%$ and $55.06\%$ for \textit{temporal-factuality} and \textit{temporal-consistency} respectively across various variants and settings.

\subsubsection{Error Analysis Cont.}
\label{appendix:error_abl}

The positive case is presented in Table \ref{tab:error_ana1} where \newmodel\ improves the temporal consistent factuality over a baseline model. It can be observed that The temporal consistency-driven factual improvement makes substantial changes both in the model temporal consistency and factuality. 

\begin{table*}[!ht]
\tiny
\centering

\begin{tabular}{@{\extracolsep{10pt}}clll}
\hline
\textbf{S.No.} & \textbf{Paraphrases} &  \textbf{Baseline Model} & \textbf{$\mathbf{\ctsrl_{Discrete}}$}\\

\hline

1&the academy award for best picture was won by the departed immediately before&	\cellcolor{LightGreen}{no country for old men}	& \cellcolor{LightGreen}{no country for old men}\\
2&the departed won the best picture academy award just before	& the reader	&\cellcolor{LightGreen}{no country for old men}\\
3&the departed won the oscar for best picture right before	& the curious case of benjamin&	 \cellcolor{LightGreen}{no country for old men}\\
4&the departed won the best picture academy award succeeded by	& the curious case of benjamin	& \cellcolor{LightGreen}{no country for old men}\\
5&movie the departed won the academy award for best picture right before& the curious case of benjamin&	 the artist\\
6&the departed won the oscar for best picture immediately prior to	& the curious case of benjamin	& \cellcolor{LightGreen}{no country for old men}\\
7&the academy award for best picture was won by the departed soon before& the curious case of benjamin&	 \cellcolor{LightGreen}{no country for old men}\\
8&the departed was awarded the academy award for best picture immediately prior to& the curious case of benjamin & the departed\\
\hline

\end{tabular}
\caption{\label{tab:error_ana1}
A comparison of output generated by a baseline model and the discrete variant of $\ctsrl$ model for the sentence completion task. Where, \textit{subject-relation} pair is "academy award for best picture - win by", \textit{key_object} is "the departed," and an expected \textit{value_object} is "no country for old men." The rest of the settings are similar as in the case of Table \ref{tab:main_res_new_model}.
}
\vspace{-3mm}
\end{table*}

\subsubsection{Probabilistic Space Analysis}
\label{appendix:prob_abl}

We conduct an exploration for the evidences of improvement in model consistency through its probabilistic space analysis. The divergence in probability distribution for the next word shall be minimal for identical intent paraphrases to generate consistent \textit{value_object} for given \textit{key_object}. Similarly, the divergence shall be wider for different intent paraphrases. We use positive and agnostic paraphrases as notations to denote paraphrases with identical intent and paraphrases with different intent. We apply random hard negative sampling while considering the agnostic paraphrases. Here is an example of a positive and agnostic paraphrases.\\

\fbox{\parbox{\linewidth-28pt}{
\textit{
\textbf{Positive paraphrases}\\
\textbf{P1:} hybrid theory was released by linkin park just before\\
\textbf{P2:} linkin park released hybrid theory immediately before\\
\textbf{Agnostic paraphrases}\\
\textbf{P1:} hybrid theory was released by linkin park just before\\
\textbf{P2:} linkin park released hybrid theory immediately after
}
}
}
The KL divergence metric is widely used to compare probabilistic distributions. We calculate the KL divergence of subsequent word's probability distribution between positive and agnostic paraphrases, respectively. The objective is to maximize the difference between these two scores. We perform this experiment on all the entities for randomly selected ten \textit{subject-relation} from test data. We randomly sample five pairs of positive and respective agnostic sentences. The scores are then averaged over a \textit{subject-relation} pair, presented in Table \ref{tab:prob_analysis}. We compare the scores between the default \llamathirteen\ model and $\ctsrl_{Discrete}$ model.

It is evident from Table \ref{tab:prob_analysis} that the \newmodel\ improves the average difference between the KL divergence scores of positive paraphrase and agnostic paraphrase by the value of $0.18$ nats. We also observe a positive change in eight \textit{subject-relation} pairs out of a total ten in the experiment. This analysis helps explain why the \newmodel\ achieves better  \textit{temporally-consistent-factuality}.

\begin{table*}[!ht]
\tiny
\centering

\begin{tabular}{@{\extracolsep{17pt}}c@{\hspace{1cm}}ccc@{\hspace{1cm}}ccc@{\hspace{1cm}}c} 
\hline

\multirow{1}{*}{\textbf{Subject-Relation}} & \multicolumn{3}{c}{\textbf{Default}} & \multicolumn{3}{c}{\textbf{$\mathbf{\ctsrl_{Discrete}}$}} & $\Delta$\\
\cline{2-4} 
\cline{5-7} 
\cline{8-8}

 \textbf{(ID)}& \textbf{PP} &  \textbf{AP} & \textbf{Diff (A)} &  \textbf{PP} &  \textbf{AP} & \textbf{Diff (B)} & \textbf{(A - B)} \\

\hline
34	&1.42&	1.52&	0.09&	1.67&	2.10&	0.43&	0.34\\
1&	5.01&	4.95&	-0.06&	5.51&	5.19&	-0.32&	-0.26\\
19&	1.99&	2.01&	0.03&	2.77&	2.91&	0.14&	0.11\\
42&	1.79&	1.72&	-0.07&	2.89&	3.36&	0.47&	0.54\\
28&	3.98&	4.07&	0.09&	4.07&	4.6&	0.54&	0.45\\
49&	1.62&	1.64&	0.02&	2.07&	2.13&	0.05&	0.03\\
36&	1.14&	1.29&	0.15&	2.96&	2.91&	-0.05&	-0.20\\
53&	3.09&	2.92&	-0.17&	2.26&	2.17&	-0.08&	0.08\\
9&	2.21&	2.22&	0.02&	3.22&	3.42&	0.20&	0.18\\
8&	2.43&	2.41&	-0.02&	2.15&	2.65&	0.50&	0.51\\

\hline

Average&2.47&	2.48&	0.01&	2.96&	3.15&0.19&	\textbf{0.18}\\

\hline

\end{tabular}
\caption{\label{tab:prob_analysis}
The results present a comparison between the default \llama[13B] model and $\ctsrl_{Discrete}$ variant (a fine-tuned \llama[13B] model) of \newmodel\ in probabilistic space. Where PP and AP are the KL divergence score between the next word's probability distribution of positive paraphrases and agnostic paraphrases, respectively, Diff is the score difference between agnostic paraphrases (AP) and positive paraphrases (PP), and $\Delta$ is the difference between the two models differentials outcome. A positive value of $\Delta$ represents that the \newmodel\ has a broader separation of divergence in agnostic paraphrases compared to positive paraphrases, reflecting a more consistent model.
}
\vspace{-3mm}
\end{table*}

\subsubsection{Significance Test for Alpha}
\label{appendix:alpha_abl}
\begin{figure*}[!ht]
 \captionsetup[subfigure]{justification=centering}
    \centering
    \subfloat[Temporal Factuality]{\includegraphics[width=0.25\textwidth]{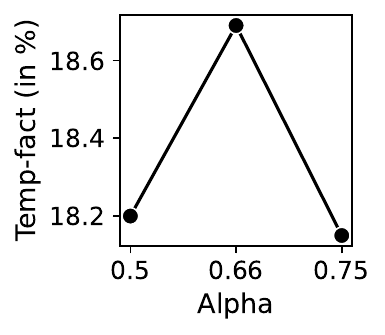}}\hspace{10mm} 
    \subfloat[Temporal Consistency]{\includegraphics[width=0.25\textwidth]{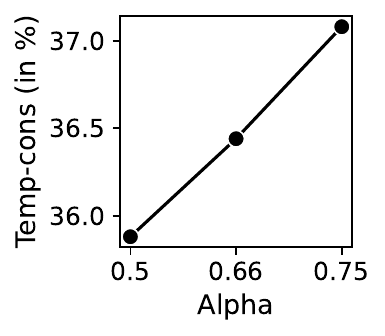}}\hspace{10mm}
    \subfloat[Temporally Consistent Factuality]{\includegraphics[width=0.24\textwidth]{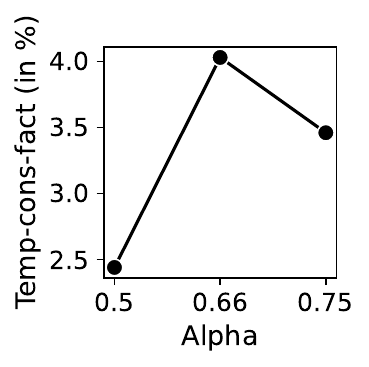}} 
    \caption{The results present the significance of a parameter alpha ($\alpha$) in the formulation of $\ctsrl_{Discrete}$ across three metrics: (a) temporal factuality, (b) temporal consistency, and (c) temporally consistent factuality respectively in percentages. The rest of the settings for this experiment are the same as in the case of Table \ref{tab:main_res_new_model}.}
    \label{fig:alpha_abl}

\end{figure*}

We carry out the experiment on significance of parameter $\alpha$ used in $\ctsrl_{Discrete}$ formulation. The experimental and data setting is the same as in section \ref{Experiments and Results:exp_result} for MT-IT + $\ctsrl_{Discrete}$ formulation. The results are presented in Figure \ref{fig:alpha_abl} for three different values of $\alpha$ (= 0.5, 0.66 and 0.75) across all three metrics \textit{temporal-factuality}, \textit{temporal-consistency} and \textit{temporally-consitent-factuality} respectively. We observe the optimal performance at $\alpha$=0.66 for \textit{temporal-factuality} and \textit{temporally-consitent-factuality}. Whereas \textit{temporal-consistency} further improves as we increase the value of $\alpha$. We have selected $\alpha$ as 0.66 for the main results presented in Table \ref{tab:main_res_new_model} based on the outcome of this ablation.

\begin{table*}[!ht]
\small
\centering
{
\begin{tabular}{@{\extracolsep{34pt}}lccc}

\hline
\textbf{Model [Parameter Size] [Quantization]} & \textbf{Temp-fact} &  \textbf{Temp-cons} &  \textbf{Temp-cons-fact} \\
\hline

\llamaseven [8-bit]& 8.10 & 24.07 & 0.00 \\
\llamathirteen [8-bit]& 18.72 & 36.88 & 4.34 \\
\llamathirty [4-bit]& 23.05 & 46.55 & 6.04 \\
\hline

\end{tabular}}
\caption{\label{tab:scalability_result}
\newmodel's ($\ctsrl_{Discrete}$) results across various parameter sizes of LLaMA.}
\vspace{-3mm}
\end{table*}

\subsubsection{Significance Test for Sample Size}
\label{appendix:samplesize_significance}
in Section \ref{Experiments and Results:exp_result}, we analyzed sampling efficiency across temporal variations in entity size and their corresponding performance, as depicted in Figure \ref{fig:temp_abl}. Furthermore, an additional experiment was undertaken to examine the relationship between the distribution of entity types and their respective performance within the test data set. Our findings indicate a notable negative correlation, quantified as -0.61, between the distribution of entity types and \textit{temporal-factuality}. Likewise, a negative correlation, valued at -0.42, was observed in relation to the distribution of entity types and \textit{temporally-consistent-factuality}. It was discerned that performance pertaining to domain-specific entity types, such as Software, Satellites, and Games, surpassed that of more general entity types, like Location and Person, with the data distribution being inversely skewed towards the latter.

\subsection{Discussion on CoTSeLF's Scalability}
\label{appendix:scalability}
In the majority of instances, the solutions that rely on parameters fine-tuning, regardless of the specific large language model (LLM) architecture involved, demonstrate scalability and adaptability to diverse models without being constrained by the number of parameters. This ablation study is performed to investigate the scalability of the proposed solution, \newmodel, specifically its capacity to scale according to the model parameter size. For this purpose, the \newmodel\ is implemented in two different \llm\ configurations, one with a reduced parameter size, \llamaseven\, and another with an increased parameter size, \llamathirty. However, it was necessary to adjust the quantization from 8-bit to 4-bit for \llamathirty\ and MT-IT's LoRA adapter both due to computational resource constraints. The experiment is conducted in same conditions as outlined in Table \ref{tab:main_res_new_model}

The findings from Table \ref{tab:scalability_result} reaffirm \newmodel's efficacy for models across a spectrum of parameter sizes, inclusive of those with large dimensions. A pronounced correlation between \newmodel's performance and the sizes of the parameters was noted. Moreover, it was observed that enhancements attributed to \newmodel\ are significant across all evaluated metrics with \llamathirty\ despite an increment in quantization level.

\subsection{Discussion on Commercial LLMs}
\label{appendix:comm_llm}

This section presents the evaluation results for two prominent commercial \llm,  \gptfour\ and \claudethree. We conducted this experiment with a limited randomly selected 100 samples from the \newdataset\ test set due to financial limitations. We set up the probe in zero-shot, where the task is to complete a sentence with the correct phrase. To this purpose, an instruction (Instruction: "complete the given sentence with the correct phrase") is provided along with the input to the model. Subsequently, we provide a concise examination of the potential for employing \newmodel\ within the realm of commercial \llm\ applications.

\subsubsection{GPT-4 Evaluation Results} 
\label{appendix:gpt4_abl}

We use 'GPT-4-0125-preview' version of \gptfour\ for this experiment. In a zero-shot setting, the \textit{temporal-factuality} stands at $36.00\%$ (Table \ref{tab:comm_llm_result}). Trained with a trillion of parameters, in absolute terms, the \textit{temporal-factuality} in \gptfour\ is not very impressive, and requires interventions such as \newmodel\ to improve it. An example of each of the positive and negative responses is mentioned in Table \ref{tab:gpt4}.

\subsubsection{Claude-3 Evaluation Results}
\label{appendix:claude3_abl}
In conducting a parallel investigation with another leading commercial LLM, \claudethree, under identical conditions, we operationalized the 'CLAUDE-3-OPUS-20240229' version for this assessment. The \textit{temporal-factuality} of \claudethree\ is noted at $20.00\%$ (Table \ref{tab:comm_llm_result}), marking a notable decline in performance relative to \gptfour.

Due to a minimal sample set, we are short of reporting either the \textit{temporally-consistent-factuality} or \textit{temporal-consistency} for this test. Metrics like temporal consistency and temporally consistent factuality require a sufficient number of pair of samples for a \textit{subject-relation} to calculate those. The other factors are temporal direction and entity pairs, which are to be considered while calculating. The 100 individual samples might be enough to gauge the preliminary assessment of temporal factuality but statistically insufficient for reporting temporally consistent factuality.

\begin{table}[!h]
\small
\centering
{
\begin{tabular}{llc}

\hline
\textbf{Model} & \textbf{Settings} &  \textbf{Temp-fact} \\

\hline
\gptfour & GPT-4-0125-preview & 36.00 \\
\claudethree & Claude-3-Opus-20240229 & 20.00\\
\hline

\end{tabular}}
\caption{\label{tab:comm_llm_result}
Temporal factuality (in \%) of two commercial \llm, \gptfour\ and \claudethree\ through the minimal sample set.}
\vspace{-3mm}
\end{table}

\begin{table*}[t]
\tiny
\centering
{
\begin{tabular}{@{\extracolsep{10pt}}clll}
\hline
\textbf{S.No.} & \textbf{Prompt} &  \textbf{Correct Answer} & \textbf{\gptfour\ Response}\\

\hline

1&Android Gingerbread was released by google immediately after android	& Froyo&\cellcolor{LightGreen}{Froyo}\\
2&Indian band Euphoria released album Sharnagat right after album & Item	&\cellcolor{LightRed}{MeHFuz}\\

\hline

\end{tabular}
}
\caption{\label{tab:gpt4}
An example of each of the positive and negative response from GPT-4 in zero-shot setting.
}
\vspace{-3mm}
\end{table*}

\subsubsection{CoTSeLF's Applicability}
\label{appendix:cotself_comm_llm}
\textbf{Applicability across commercial \llm.} Commercial model's such as \gptfour\ and \claudethree's \textit{temporal-factuality} remains suboptimal, thus necessitating strategies like CoTSeLF or similar enhancements to augment its  \textit{temporal-factuality}. \newmodel\ advances an RL-based fine-tuning methodology that refines the model's parameters via a custom-made fine-tuning procedure. Almost universally, advancements in parameter fine-tuning methodologies are contingent upon access to the model's architecture. GPT-4 and Claude-3, by restricting direct architectural access, necessitate that such methodologies could demonstrate their effectiveness only on open-source models. Thus, we could not assess \newmodel's efficacy on \gptfour\ or \claudethree\ models due to the inaccessibility of its architecture for open fine-tuning.

Considering that \gptfour\ adhered to a decoder-based transformer architecture (up to GPT-2) before the later versions transitioned to a commercial model, there is an intense anticipation that \newmodel, particularly \ctsrl, will serve as an efficacious approach to augmenting the closed-source model's capabilities in improving \textit{temporally-consistent-factuality} as in case of various versions of LLaMA model under experimentation.

An alternative approach involves examining methods that do not necessitate access to the model's architecture for parameter updates, acknowledging that typically, strategies not involving parameter modifications exhibit suboptimal performance compared to those that do, like \newmodel. Future research should focus on enhancing the \textit{temporally-consistent factuality} of commercial \llm\ through techniques that might bypass the need for gradient updates.

\textbf{Applicability across domains.} The establishment of consistent temporal reasoning capabilities is poised to profoundly influence their applicability across a spectrum of complex fields, including healthcare and the legal sector. Specifically, within the legal domain, the temporally reliant nature of case precedents and statutory amendments demands the reliable and precise retrieval of information. LLMs must avoid offering divergent interpretations of precedents for specific case types to prevent severe repercussions and undermine trust in the judicial system. The precedence of rules, articles, and cases, being temporally bound, demands their consistent retrieval when employing LLMs in legal settings. Similarly, in healthcare, the prognostication of diseases is time-sensitive, mandating consistent and precise forecasts for diagnosing and formulating treatment plans, leading to enhanced trust in the usage of AI-based medical solutions. Techniques like CoTSeLF could be tailored to meet such requirements.

\subsection{Hyperparameters}

All codes were composed utilizing PyTorch. We utilized the Huggingface\footnote{\url{https://huggingface.co/}} repository for stacking the open-source \llm. The implementation of RL is carried out through trlX, a python-based library \cite{trlx-library}. A PEFT-based method LoRA with 8-bit quantization is used for both instruction-tuned and PPO-based RL models. We apply minimal pre-processing on generated outputs from \llm\, such as filtering of articles or special characters (at max up to one word) before retrieving the \textit{value_object}. The maximum sequence length of $256$ is used across all experiments.  An NVIDIA A100 GPU (80 GB)\footnote{\url{https://www.nvidia.com/en-in/data-center/a100/}} was used to train the model.

\end{document}